\documentclass[11pt]{article}
\RequirePackage[OT1]{fontenc}
\RequirePackage{amsthm,amsmath,natbib}
\RequirePackage[colorlinks]{hyperref}
\usepackage{amsfonts}
\usepackage{amssymb}
\usepackage{amstext}
\usepackage{parskip}
\usepackage{fullpage}
\usepackage[ruled,linesnumbered,vlined]{algorithm2e}
\usepackage{booktabs}
\usepackage{multirow}
\usepackage{array}
\usepackage{url}
\usepackage{enumitem}

\usepackage{graphicx}
\usepackage[footnotesize]{subfigure}

\DeclareFixedFont{\auacc}{OT1}{phv}{m}{n}{12}   

\def\math{${}^\diamond$}
\def\stat{${}^\star$}

\providecommand{\keywords}[1]{\textbf{\textit{Keywords---}} #1}

\newenvironment{figrow}%
{%
\centering\addtocounter{figure}{1}
\begin{enumerate}[%
itemsep=2pt,parsep=0em,
label={(\alph*)},
ref={\thefigure.(\alph*)}
]}%
{\end{enumerate}\addtocounter{figure}{-1}}

\begin{document}
\title{Spatial Projection of Multiple Climate Variables \\Using Hierarchical Multitask Learning
\footnote{Paper published in the 31st AAAI Conference on Artificial Intelligence (AAAI'17). San Francisco, CA. 2017.}}

\author{Andr\'e R. Gon\c{c}alves\stat\math,\;\;
 Arindam Banerjee\stat,\;\; Fernando J. Von Zuben\math\;
\\ {\math}School of Electrical and Computer Enginnering, University of Campinas, Brazil\\
{\stat}Department of Computer Science, University of Minnesota, Twin Cities, USA\\
\{andre@cs.umn.edu, banerjee@cs.umn.edu, vonzuben@dca.fee.unicamp.br\} }
\date{}
\maketitle
	\begin{abstract}
Future projection of climate is typically obtained by combining outputs from multiple Earth System Models (ESMs) for several climate variables such as temperature and precipitation. While IPCC has traditionally used a simple model output average, recent work has illustrated potential advantages of using a multitask learning (MTL) framework for projections of individual climate variables. In this paper we introduce a framework for hierarchical multitask learning (HMTL) with two levels of tasks such that each {\em super-task}, i.e., task at the top level, is itself a multitask learning problem over \emph{sub-tasks}. For climate projections, each super-task focuses on projections of specific climate variables spatially using an MTL formulation. For the proposed HMTL approach, a group lasso regularization is added to couple parameters across the super-tasks, which in the climate context helps exploit relationships among the behavior of different climate variables at a given spatial location. We show that some recent works on MTL based on learning task dependency structures can be viewed as special cases of HMTL. Experiments on synthetic and real climate data show that HMTL produces better results than decoupled MTL methods applied separately on the super-tasks and HMTL significantly outperforms baselines for climate projection.
	\end{abstract}
	\section{Introduction}
	
	Future projections of climate variables such as temperature, precipitation, and pressure are usually produced by physics-based models known as Earth System Models (ESMs). ESMs consist of four components and their interactions, viz. atmosphere, oceans, land, and sea ice \citep{Tebaldi2007,IPCC5Report}. Climate projections generated from ESMs form the basis for understanding and inferring future climate change, global warming, greenhouse gas concentration and its impact on Earth systems and other complex phenomena such as El Ni\~no Southern Oscillation (ENSO). ENSO, for instance, has a global impact, ranging from droughts in Australia and northeast Brazil to heavy rains over Malaysia, the Philippines, and Indonesia \citep{IPCC5Report}. Then, producing accurate projections of climate variables is essential to anticipate extreme events.
		
	Many ESMs have been developed by climate research institutes. A single and possibly more robust projection can be built as a combination (ensemble) of multiple ESMs simulations \citep{Tebaldi2007,Mcquade2013,Sanderson2015}.
	
	Recently, the problem of constructing ESMs ensemble was approached from a multitask learning (MTL) perspective \citep{Goncalves2014}, where building an ESMs ensemble for each geographical location was viewed as a learning task. The joint estimation of the ESM ensemble produced more accurate projections than when independent estimation was performed for each location. The MTL method was able to capture the relationship among geographical locations (tasks) and used it to guide information sharing among tasks.
	
	Modeling task relationship in multitask learning has been the focus of recent research  \citep{Zhang-Schneider2010,Zhang2012,Yang2013,Goncalves2016}. This is a fundamental step to promote information sharing only among related tasks, while avoiding the unrelated ones. Besides estimating task specific parameters ($\Theta$), the task dependency structure ($\Omega$) is also estimated from the data. The latter is usually estimated from the former, i.e., task dependency is defined based on the relation of the task parameters.  Two tasks are said to be related if their parameters are related in some sense.
	
	 Inconsistently estimated task dependency structure in MTL can misguide information sharing and, hence,  can be harmful to the MTL method performance. The problem of estimating statistical dependency structure of a set of random variables is known as structure learning  \citep{Meinshausen2006}. Existing methods for the problem guarantee to recover the true underlying dependence structure given a sufficient amount of data samples. In the MTL case \citep{Goncalves2016}, the random variables are tasks parameters and, depending on the ratio between dimensionality and the number of tasks, the amount of data samples may not be sufficient.
	
	In this paper, we introduce the \emph{Hierarchical Multitask Learning} (HMTL) method that  jointly learn multiple tasks by letting each task be, by itself, a MTL problem. The associated hierarchical multitask learning problem is then said to be composed of \emph{super-tasks} and the tasks involved in each super-task as \emph{sub-tasks}. Our formulation is motivated by the problem of constructing ESM ensembles for multiple climate variables, with multiple geolocations for each variable. The problem of obtaining ESMs weights for all regions for a certain climate variable is a super-task. 
	
	The paper is organized as follows. We first discuss existing MTL methods with task dependency estimation, on which our proposed ideas are built; then, we briefly talk about the climate projection problem that  motived our work. In the sequel, we present our proposed framework, perform experiments and analyze the results. Concluding remarks and next research steps complete the paper. \\

	\noindent \emph{Notation}. Let $T$ be the number of super-tasks, $d$ the problem dimension, and $n_{(t,k)}$ the number of samples for the $(t,k)$-th sub-task. For the purposes of the current paper, we assume that all super-tasks have $m$ sub-tasks and all sub-tasks have problem dimension $d$. $X^{(t,k)}\in \mathbb{R}^{n_{(t,k)}\times d}$ and  $\mathbf{y}^{(t,k)}\in \mathbb{R}^{n_{(t,k)}\times 1}$ are the input and output data for the $k$-th sub-task of the $t$-th super-task. $\Theta^{(t)} \in \mathbb{R}^{d\times m}$ is the matrix whose columns are the set of weights for all sub-tasks for the $t$-th super-task, that is, $\Theta^{(t)} =[\boldsymbol{\theta}^{(t,1)},...,\boldsymbol{\theta}^{(t,m)}]$. For the ease of exposition, we represent $\{X\} = X^{(t,k)}$ and $\{Y\} = \mathbf{y}^{(t,k)}, k=1,...,m_t; t=1,...,T$. For the weight and precision matrices, $\{\Theta\} = \Theta^{(t)}$ and $\{\Omega\}=\Omega^{(t)}, \forall t=1,...,T$. Identity matrix of dimension $m\times m$ is denoted by $\mathbf{I}_{m}$. $\mathcal{U}(a,b)$ is an uniform probability distribution in the range [a,b].

	\section{Multitask Learning with Task Dependence Estimation}
	\label{sec:related_work_hmtl}
	
Explicitly modeling task dependencies has been made by means of Bayesian models. Features across tasks (rows of the parameter matrix $\Theta$) were assumed to be drawn from a multivariate Gaussian distribution. Task relationship is then encoded in the inverse of the covariance matrix $\Sigma^{-1}=\Omega$, also known as \emph{precision matrix}. Sparsity is desired in such matrix, as zero entries of the precision matrix indicate conditional independence between the corresponding two random variables (tasks) \citep{Friedman2008}. The associated learning problem \eqref{eq:mtl_struct_hmtl} consists of jointly estimating the task parameters $\Theta$ and the precision matrix $\Omega$, which is done by an alternating optimization procedure.
	\begin{equation}
	\begin{aligned}
	& \underset{\Theta,\Omega}{\text{min}}
	& &\sum_{k=1}^{m} L(X_k,\mathbf{y}_k, \Theta) - \log |\Omega| + \lambda_0\textrm{tr}(\Theta\Omega\Theta^{\top}) + R(\Theta,\Omega)\\
	& \text{s.t.}
	& &\Omega \succeq 0.
	\label{eq:mtl_struct_hmtl}	
	\end{aligned}
	\end{equation}
	Note that in \eqref{eq:mtl_struct_hmtl} the regularization penalty $R(\Theta,\Omega)$ is a general penalization function that will be discussed later in the paper. A solution for \eqref{eq:mtl_struct_hmtl} alternates between the following two steps until a stopping criterion is met:
	\begin{description}
		\item[Step 1] Estimate $\Theta$ from current estimation of $\Omega$;
		\item[Step 2] Estimate $\Omega$ from updated parameters $\Theta$.
	\end{description}
	Setting initial $\Omega$ to identity matrix, i.e., all tasks are independent at the beginning, is usually a suitable start.
	
	In Step 1, task dependency information is incorporated into the joint cost function through the trace term penalty $\textrm{tr}(\Theta\Omega\Theta^{\top})$. It helps to promote information exchange among tasks. The problem associated with Step 2, known as \emph{sparse inverse covariance selection problem} \citep{Friedman2008}, seeks to find some sparsity pattern in the precision matrix. Experimental analysis have shown that these approaches usually outperform MTL with pre-defined task dependency structure for a variety of problems \citep{Zhang-Schneider2010,Goncalves2014}.
	
	\section{Mathematical Formulation of ESMs Climate Projection}
	\label{sec:climate_projection}
	
	A common projection method is to perform the combination of multiple ESMs in a least square sense, that is, to estimate a set of weights for the ESMs based on past observations. 
	
	For a given location $k$ the predicted climate variable (temperature, for example) for a certain timestamp $i$ (expected mean temperature for a certain month/year, for example) is given by:
	\begin{equation}
	\hat{y}_k^i = \langle\mathbf{x}_k^i,\boldsymbol{\theta}_k\rangle + \epsilon_k^i
	\label{eq:least_square_hmtl}
	\end{equation}
	where $\mathbf{x}_k^i$ is the set of values predicted by the ESMs for the $k$-th location in the timestamp $i$, $\boldsymbol{\theta}_k$ is the weight vector of each ESM for the $k$-th location, and $\epsilon_k^i$ is a residual. The weight vector $\boldsymbol{\theta}_k$ is estimated from a training data. The combined estimate $\hat{y}_k^i$ is then used as a more robust prediction of temperature for the $k$-th location in a certain month/year in the future.
	
	ESMs weights ($\boldsymbol{\theta_k}$) are defined for each geographical location and it possibly varies for different locations. It is possible that some ESMs are more accurate for some regions/climate than others and the difference between weights of two locations will reflect this behavior. In summary, the ESMs ensemble consists of solving a least square problem for each geographical location.
	
	The ESMs weights may vary for the prediction of different climate variables, such as precipitation, temperature, and pressure.  Then, solving an MTL problem  for each climate variable is required.  In this paper, we propose to simultaneously tackle multiple MTL problems through a hierarchical (two-level) MTL formulation.

	\section{The HMTL Formulation}
	\label{sec:hmtl}
	
	The HMTL formulation seeks to minimize the following cost function $C(\boldsymbol{\Gamma})$ with $\boldsymbol{\Gamma} =\{\{\Theta\},\{\Omega\},\lambda_0\}$:
	\begin{align}
	\mathcal{L}(\boldsymbol{\Gamma}) & =  \sum_{t=1}^{T} \left(\sum_{k=1}^{m_t} L \left(X^{(t,k)}\boldsymbol{\theta}^{(t,k)},\mathbf{y}^{(t,k)} \right) \right. \nonumber \\ & \;\left. -\log| \Omega^{(t)}| + \lambda_0\textrm{tr}\left(S^{(t)}\Omega^{(t)}\right)\right) + \mathcal{R}(\{\Omega\})
	\label{eq:jmssl_formulation}
	\end{align}
	\noindent where $R(\{\Omega\})$ is a regularization term over the precision matrices, $S^{(t)}$ is the sample covariance matrix of the task parameters for the $t$-th super-task. For simplicity, we dropped the $\ell_1$-penalization on the weight matrix $\Theta$ as is often done in MTL \citep{Zhang-Schneider2010,Yang2013,Goncalves2016}. However, it can be added, if one desires sparse $\Theta$, with minor changes in the algorithm. All super-tasks are assumed to have the same number of sub-tasks. Squared loss was used as loss function.
	
	The formulation (\ref{eq:jmssl_formulation}) is a penalized cumulative cost function of the form \eqref{eq:mtl_struct_hmtl} for several multitask learning problems. The penalty function $R(\{\Omega\})$ is to favor common structural sparseness across the precision matrices. Here, we focus on the group lasso penalty \citep{Yuan2006}, which we denote by $R_G$, and is defined as
		\begin{equation}
		R_{G}(\{\Omega\}) = \lambda_1\sum_{t=1}^{T}\sum_{k\neq j}|\Omega_{kj}^{(t)}| + \lambda_2\sum_{k\neq j}\sqrt{\sum_{t=1}^{T} \Omega_{kj}^{(t)^2}}
		\end{equation}
		where $\lambda_1$ and $\lambda_2$ are two nonnegative tuning parameters. The first penalty term is an $\ell_1$-penalization of the off-diagonal elements, so that non-structured sparsity in the precision matrices is enforced. The larger the value of $\lambda_1$, the sparser the precision matrices. The second term of the group sparsity penalty encourages the precision matrices for different super-tasks to have the same sparsity pattern. Group lasso does not impose any restriction on the degree of similarity of the non-zero entries of the precision matrices.
		
	\begin{figure*}[htb]
		\raisebox{-0.8\height}{\includegraphics[scale=0.95]{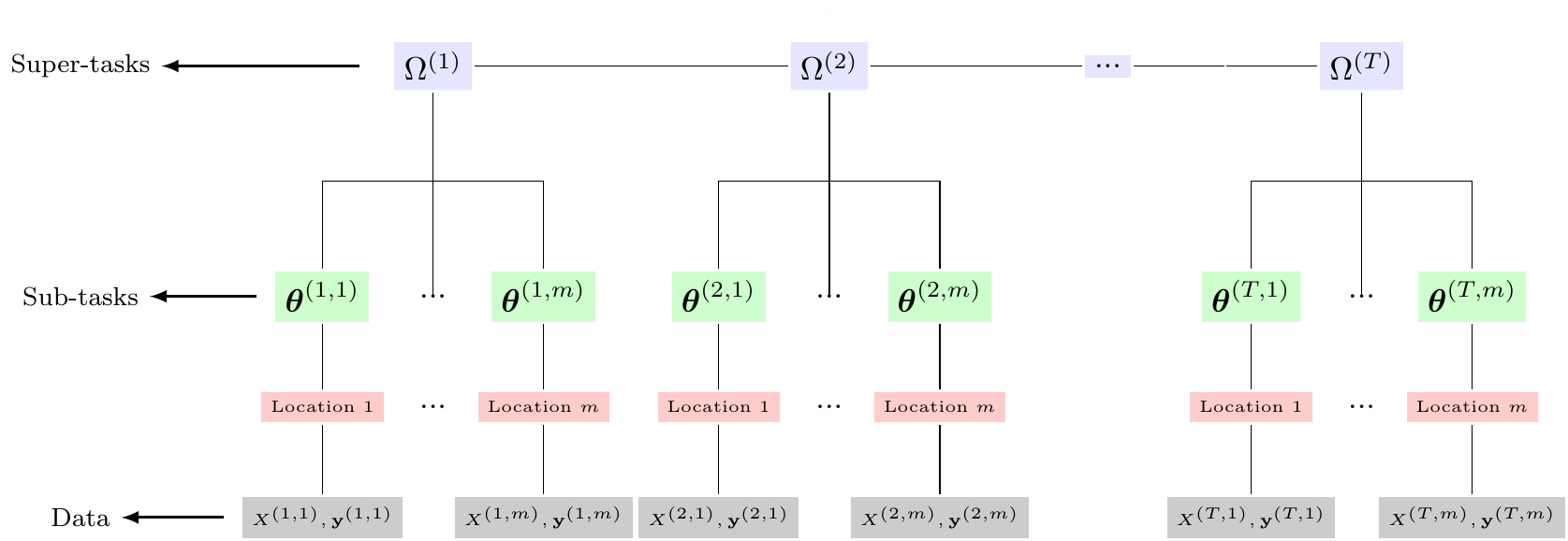}}
		\caption{Hierarchy of tasks and their connection to the climate problem. Each super-task is a multitask learning problem for a certain climate variable, while sub-tasks are least square regressors for each geographical location.}
		\label{fig:hierarchical_diagram_hmtl}
	\end{figure*}

	Setting $\lambda_2$ to zero, super-tasks are decoupled into independent MTL formulations. Then, $\lambda_2$ can be seen as a coupling parameter, as larger values push the super-tasks to be coupled, so that different $\Omega^{(t)}$ have similar sparsity patterns. On the other hand, lower (or zero) values lead to decoupled super-tasks, recovering existing MTL formulations, such as in \citep{Goncalves2014}. 
	
	Figure~\ref{fig:hierarchical_diagram_hmtl} shows the hierarchy of tasks for the projection of multiple climate variables. At the level of super-tasks,  group lasso regularization encourages precision matrices to have a similar sparseness pattern. The learned precision matrices are consequently used to control with whom each sub-task will share information.

 The correspondence between the variables in the HMTL formulation and the elements in the climate problem is shown in Table \ref{tab:correspondence_mtl_climate}. 
	
	\begin{table}[htb]
	\centering
		\small
				\caption{Correspondence between HMTL variables and the components in the joint ESMs ensemble for the multiple climate variables problem.}
			\label{tab:correspondence_mtl_climate}
		\begin{tabular}{p{1.5cm}p{5cm}p{5cm}}
			\toprule
			\textbf{Variable} & \textbf{HMTL meaning} & \textbf{Climate meaning} \\ \toprule
			$T$ & \# of super-tasks & \# of climate variables \\ \hline
			$m$ & \# of sub-tasks in the $t$-th super-task & \# of locations (equal for all climate variables) \\ \hline
			$X^{(t,k)}$ & data input for the $k$-th sub-task in the $t$-th super-task & ESMs predictions for the $t$-th climate variable in the $k$-th location \\ \hline
			$\mathbf{y}^{(t,k)}$ & data output for the $k$-th sub-task in the $t$-th super-task & observed values of the $t$-th climate variable in the $k$-th location \\ \hline
			$\boldsymbol{\theta}^{(t,k)}$ & linear regression parameters for the $k$-th sub-task of the $t$-th super-task &  ESMs weights for the $t$-th climate variable in the $k$-th location \\ \hline
			$\Omega^{(t)}$ & precision matrix for the $t$-th super-task & dependence among the ESMs weights for all locations for the $t$-th climate variable
			\\ \bottomrule
		\end{tabular}
	\end{table}
	
	\subsection{Optimization}
	
	Optimization problem~\eqref{eq:jmssl_formulation} is not jointly convex on $\{\Theta\}$ and $\{\Omega\}$, particularly due to the \emph{trace} term which involves both variables. We then use an alternating minimization, in which $\{\Theta\}$ is held fix and optimize for $\{\Omega\}$ (we call it $\Omega$-step), and similarly fix $\{\Omega\}$ and optimize for $\{\Theta\}$ (we call it $\Theta$-step). Both steps now consist of convex problems, for which efficient methods have been proposed. In the experiments, 20 to 30 iterations were required for convergence.
	
	\subsubsection{Solving $\Theta$-step} The convex problem associated with this step is defined as
	\begin{equation}
	\begin{aligned}
	& \underset{\{\Theta\}}{\text{min}}
	& &  \sum_{t=1}^{T}   \sum_{k=1}^{m_k} L\left(X^{(t,k)}\boldsymbol{\theta}^{(t,k)},\mathbf{y}^{(t,k)} \right)+ \lambda_0\textrm{tr}\left(S^{(t)}\Omega^{(t)}\right).
	\end{aligned}
	\label{eq:jmssl_step1}
	\end{equation}
	
	Considering the squared loss function, $\Theta$-step consists of two quadratic terms, as $\{\Omega\}$ are positive semidefinite matrices. Note that the optimization for each super-task weight matrix $\Theta^{(t)}$ are independent and can be performed in parallel. We used the L-BFGS \citep{Liu1989} method in the experiments.
	
	\subsubsection{Solving $\Omega$-step}
	The $\Omega$-step is to solve the following optimization problem
	\begin{equation}
	\begin{aligned}
	& \underset{\{\Omega\}}{\text{min}}
	& &   \sum_{t=1}^{T} \Big(-\log|\Omega^{(t)}| + \lambda_0\textrm{tr}(S^{(t)}\Omega^{(t)}) \Big) + R_{G}(\{\Omega\})\\
	& \text{s.t.}
	& & \Omega^{(t)} \succeq 0, \;\;\forall t=1,...,T.
	\end{aligned}
	\label{eq:jmssl_step2}
	\end{equation}
	This step corresponds to the problem of joint learning multiple Gaussian graphical models and has been recently studied \citep{Honorio2010,Danaher2014,Mohan2014}. These formulations seek to minimize the penalized joint negative log likelihood in the form of (\ref{eq:jmssl_step2}) and they basically differ in the penalization term $R(\{\Omega\})$. Researchers have shown that the graphical models jointly estimated were able to increase the number of edges correctly identified while reducing the number of edges incorrectly identified, when compared to those independently estimated. An alternating direction method of multipliers (ADMM) proposed in \citep{Danaher2014} was used to solve problem (\ref{eq:jmssl_step2}).
		
	Algorithm \eqref{alg:hmtl} presents the pseudo-code for the proposed HMTL algorithm.
	
	\begin{algorithm}
		\DontPrintSemicolon
		\KwData{$\{\mathbf{X}\}$, $\{\mathbf{Y}\}$.} 
		\KwIn{$\lambda_0>0$, $\lambda_1>0$ and $\lambda_2>0$. } 
		\KwResult{$\{\Theta\}$, $\{\Omega\}$. } 
		\Begin{
			$\Omega^{(t)}=\mathbf{I}_{m_t}, \forall t=1,...,T.$\\
			$\Theta^{(t)}=\mathcal{U}(-0.5,0.5), \forall t=1,...,T.$\\
			\Repeat{stopping condition met}{
				Update $\{\Theta\}$ by solving \eqref{eq:jmssl_step1}; \\
				Update $\{\Omega\}$ by solving \eqref{eq:jmssl_step2}; 
			}
		}
		\caption{HMTL algorithm.}
		\label{alg:hmtl}
	\end{algorithm}
	
	\section{Related Works}
	
	A lattice graph is used to represent the regional relationship in \citep{Karthik2013}, where immediate neighbor locations are assumed to have similar ESMs weights. Weights for all geolocations are estimated jointly (in a least square sense) with graph Laplacian regularization to encourage spatial smoothness.
	
	In \citep{Mcquade2013} the online ESMs ensemble problem is tackled by using a lattice Markov Random Field. The state of each hidden variable, which is associated with a geographical location,  is the identity of the best ESM for that specific location. The marginal probabilities of the hidden variables act as the weights of the ensemble. Hence, ESMs with a higher probability of being the best has a larger weight in the ensemble.  At each time step marginal probabilities are updated based on the performance of the ESMs in the previous time step via the loopy belief propagation algorithm.
	
	Differently, \citep{Goncalves2014} do not assume any fixed dependence graph, but, instead, estimate it 
	through a multitask learning model that makes use of a sparse Gaussian Markov Random Field (GMRF) to capture the relationship of ESMs weights among locations. ESM weights for all locations and the GMRF are jointly estimated via an alternating minimization scheme.

	Our work differs from the existing research as it (1) leverages information not only from immediate neighbors but also from any related geographical locations; (2) allows two levels of information sharing: ESMs weights and precision matrices that encodes the relationship of locations; and (3) handles the projection of multiple climate variables (exploring their resemblance) simultaneously.

	\section{Experiments}
	\label{sec:experiments}
	In this section we present experiments to compare the proposed HMTL with existing methods in the literature for both synthetic and real climate data.

	\subsection{Synthetic Data}
	
		We first generated a synthetic dataset to assess the performance of HMTL over traditional MTL methods. For comparison, we used the MTL method proposed in \citep{Goncalves2014,Goncalves2016}, called MSSL, that has shown to be competitive with existing MTL algorithms including \citep{Zhang2012,Kumar2012,Kang2011}. We also seek to investigate the effect of the increase in the number of super-tasks. For this analysis we generated 7 super-tasks containing 15 sub-tasks each, with dimension of 50. For each sub-task 100 data samples were generated. Inverse covariance matrices $\Omega^{(t)}, t=1,...,T$ were drawn from a Wishart distribution with a scale matrix $\Lambda$ and $n=10$ degrees of freedom. Scale matrix $\Lambda$ was designed to reflect a structure containing three groups of related variables. As sampled precision matrices are also likely to present the group structure, jointly sparse precision matrices are suitable to capture such pattern, which is precisely what HMTL formulation assumes.
		
		Given $\Omega^{(t)}$, the sub-tasks parameters $\Theta^{(t)}$ were constructed as: $\Theta^{(t)}(j,:) = \mathcal{N}(0,\Omega^{(t)}),\; j=1,..,d$; and, finally,  the design matrix $X^{(t,k)}$ were sampled from $\mathcal{N}(0,I)$, and $Y^{(t,k)} = X^{(t,k)}\Theta^{(t)}+\epsilon^{(t,k)}$, where $\epsilon^{(t,k)}\sim \mathcal{N}(0,0.1)$, $\forall k=1,..,m;\;t=1,..,T.$
			
		 Ten independent runs for both HMTL and MSSL were carried out. Each run with a different random train/test data split. Table~\ref{tab:relative_gain} shows the relative improvement\footnote{Relative improvement is given by the difference between MSSL and HMTL performance (RMSE) divided by MSSL performance  as percents (\%).} of HMTL over MSSL for distinct number of super-tasks.  Two scenarios were tested: using 50 and 30 samples for training and the remaining for test (50 and 70, respectively).  Penalization parameters ($\lambda$'s) were chosen by cross-validation on the training set.
	\begin{table}[!htb]
		\centering
			\caption{Average relative improvement (in \%) of HMTL over MSSL for the synthetic dataset. HMTL produced lower RMSE as the number of super-tasks increase.}
					\label{tab:relative_gain}
		\begin{tabular}{c|cccccc}
				\hline	
				Training & \multicolumn{6}{c}{\# of super-tasks} \\	\cline{2-7}
				samples & 2 & 3 & 4 & 5 & 6 & 7 \\	\hline	
				50 &  2.2  &  4.0 &   4.9  &  5.5 &   5.2 &   5.8 \\
				30 & 22.3  & 23.4 &  25.1  & 25.7 &  24.2 &  24.3 \\ \hline
		\end{tabular}
	\end{table}
	
	From Table~\ref{tab:relative_gain}, we observe that HMTL has a continuous increase on the relative improvement compared to independently running MSSL, as the number of super-tasks grows. From the sixth super-task on, the relative improvement shows a tendency to stabilize. Such behavior is also observed in the scenario with only 30 training samples. It is worth noting that the smaller a set of training data (30 samples) the sharper the improvements obtained by HMTL over MSSL.
	
	\subsection{Climate Data}
	
	We collected monthly land temperature and precipitation data of 32 CMIP5 ESMs \citep{tasm12}, from 1901 to 2000, in South America. Observed data provided by \citep{Willmott2001} was used. ESMs predictions and observed values from 250 locations in South America (distributed in a grid shape) were considered. From the HMTL perspective, the problem involves: two super-tasks, 250 sub-tasks (per super-task) with dimensionality of 32.
	
	In the climate domain, it is common to work with the relative measure of the climate variable to a value of reference, which is obtained from past information. In our experiments, we directly work on the raw data (not detrended). We investigate the performance of the algorithm in both seasonal and annual time scales, with focus on winter and summer. All ESMs and observed data are in the same time and spatial resolution. Temperature is in degree Celsius and precipitation in cm.

	\subsubsection{Experimental Setup:}
	
	Based on climate data from a certain past (training) period, model parameters are estimated and the inference  produces its projections for the future (test). Clearly, the length of the training period affects the performance of the algorithm. A moving window of 20, 30 and 50 years were used for training and the next 10 years for test. The performance is measured in terms of root-mean-squared error (RMSE).
	
	Seasonality strongly affect climate data analysis. Winter and summer precipitation patterns, for example, are distinct. Also, by looking at seasonal data, it becomes easier to identify anomalous patterns, possibly useful to characterize climate phenomena as El Ni\~no. We extracted summer and winter data and performed climate variable projection specifically for these seasons.

\begin{table*}[!ht]
   \footnotesize
	\centering
	\begin{tabular}{cccccccc}
		\hline
		& \# of training     & Best-ESM   &	 OLS 	     & S$^2$M$^2$R   &     MMA     & MSSL & HMTL \\ \cline{3-8}
		& years &\multicolumn{6}{c}{\textbf{Precipitation}} \\ \hline
		& 20 &  7.88 (0.44)  & 9.08 (0.54)  & 7.33 (0.68)  & 8.95 (0.27)  & 7.16 (0.43) &\textbf{ 6.48 (0.34)$^{**}$} \\
		Summer  & 30 & 7.95 (0.55)   & 7.87 (0.63)  & 7.39 (0.86) &  8.96 (0.26)  & 6.86 (0.48) & \textbf{6.37 (0.29)$^*$} \\
		& 50 &  8.30 (0.71)  & 7.84 (1.13) &  7.86 (1.12) &  9.03 (0.30) & 6.89 (0.55) & \textbf{6.42 (0.33)}\\ \hline
		
		& 20 & 4.83 (0.26)  & 5.62 (0.30) &  4.58 (0.39) &  5.44 (0.24) & 3.98 (0.21)  & \textbf{3.83 (0.22) }\\
		Winter	  & 30 & 4.86 (0.29)  & 4.83 (0.27) &  4.68 (0.38)  & 5.41 (0.25) & 3.94 (0.17) & \textbf{3.80 (0.21)$^{*}$} \\
		& 50 & 4.92 (0.38) &  4.64 (0.63) &  4.77 (0.52) &  5.33 (0.18)  & 3.84 (0.21) & \textbf{3.70 (0.20)} \\ \hline
		
		& 20 & 7.38 (0.17)  & 6.03 (0.65) & 6.49 (0.49) & 7.78 (0.14) & 5.79 (0.16) & \textbf{5.70 (0.16)} \\
		Year       & 30 & 7.41 (0.18) &  6.21 (0.80)  & 6.57 (0.61)  & 7.76 (0.14) & 5.72 (0.16) &\textbf{ 5.66 (0.18)} \\
		& 50 & 7.47 (0.26)  & 6.56 (1.07)  & 6.87 (0.80) &  7.73 (0.14) & 5.69 (0.23) & \textbf{5.61 (0.22)} \\ \hline
		\multicolumn{8}{c}{\textbf{Temperature}}\\ \hline
		& 20 & 1.39 (0.23)  & 1.22 (0.10) &  0.95 (0.13) &  1.95 (0.02) & 0.82 (0.08) & \textbf{0.81 (0.01)} \\
		Summer & 30 & 1.47 (0.30)  & 1.21 (0.15)  & 1.09 (0.17)  & 1.96 (0.01)& 0.84 (0.07) & \textbf{0.80 (0.01)} \\
		& 50 & 1.63 (0.35)  & 1.40 (0.19)  & 1.36 (0.20)  & 1.98 (0.01) & 0.88 (0.05) & \textbf{0.83 (0.01})$^{*}$\\ \hline
		
		& 20 & 1.58 (0.19) &  1.48 (0.08)  & 1.18 (0.12) &  2.08 (0.01) & 1.03 (0.04) & \textbf{1.02 (0.03)} \\
		Winter   & 30 & 1.64 (0.26)  & 1.40 (0.13) &  1.27 (0.16) &  2.09 (0.01) & 1.01 (0.04) & \textbf{0.99 (0.03)} \\
		& 50 &  1.77 (0.31)  & 1.55 (0.17)  & 1.51 (0.18) &  2.08 (0.01) & 1.04 (0.02) & \textbf{0.98 (0.03)$^{**}$} \\  \hline
		
		& 20 & 1.64 (0.18) &  1.10 (0.13)  & 1.13 (0.12) &  2.11 (0.01) & 1.00 (0.04) & \textbf{0.91 (0.02)$^{**}$} \\
		Year 	  & 30 &  1.70 (0.24) &  1.20 (0.17)  & 1.24 (0.17) &  2.12 (0.01) & 1.00 (0.04) & \textbf{0.91 (0.02)$^{**}$} \\
		& 50 & 1.83 (0.28) &  1.47 (0.21)  & 1.50 (0.20) &  2.12 (0.01) & 1.01 (0.03) & \textbf{0.91 (0.02)$^{**}$}\\ \hline 			
	\end{tabular}
	\caption{Precipitation and Temperature: Average and standard deviation RMSE for all scenarios. HMTL presented the best results. Two-sampled t-test was performed and statistically significant differences between HMTL and the second best method at a level of $p$ < 0.05(*) and $p$ < 0.01(**) are highlighted.}
	\label{tab:rmse_precip}
\end{table*}

	Five baseline algorithms were considered:
	
	\begin{enumerate}
		\item \textbf{multi-model average} (MMA): set equal weights for all ESMs. This is currently performed by IPCC \citep{IPCC5Report};
		\item \textbf{best-ESM} in training phase: it is not an ensemble, but a single best ESM in terms of mean squared error;
		\item \textbf{ordinary least square} (OLS): perform independent OLS for each location and climate variable;
		\item \textbf{S$^2$M$^2$R} \citep{Karthik2013}: can be seen as an MTL method with pre-defined location dependence given by the graph Laplacian. It incorporates spatial smoothing on ESMs weights.
		\item \textbf{MSSL} \citep{Goncalves2014}: run MSSL for each climate variable projection independently. The parameter-based version (\textit{p}-MSSL) was used.
	\end{enumerate}
	
	All the penalization parameters of the methods ($\lambda$'s in MSSL and HMTL) were chosen by cross-validation. From the training set, we selected the first 80\% for training and the next 20\% for validation. The best values in the validation set were selected. For example, in the scenario with 20 years of measurements for training, we took the first 16 years to really train the model, and the next 4 years to analyze the performance of the method using a specific setting  of $\lambda$'s. Using this protocol, the selected parameter values were: S$^2$M$^2$R used $\lambda=1000$; MSSL $\lambda_0=0.1$	and $\lambda_1=0.1$; and HMTL $\lambda_0=0.1$,	 $\lambda_1=0.0002$,	$\lambda_2=0.01$.  
	
	\subsubsection{Results:}
	
	Table~\ref{tab:rmse_precip} shows the RMSE of the projections produced by the algorithms and the observed values, for precipitation and temperature. First, we note that simply assigning equal weights to all ESMs does not seem to exploit the potential of ensemble methods. MMA (as used by IPCC, \citep{IPCC5Report}) presented the largest RMSE among the algorithms for the majority of periods (\emph{summer}, \emph{winter} and \emph{year}) and number of years for training. 	Second, the MTL methods, MSSL and HMTL, clearly outperform the baseline methods. S$^2$M$^2$R does not always produce better projections than OLS. In fact, it is slightly worse for the \emph{year} dataset. As expected, the assumption of spatial neighborhood dependence does not seem to hold for all climate variables.

	HMTL presented better results than performing MSSL for precipitation and temperature independently in many situations. HMTL was able to significantly reduce RMSE in  \emph{summer} precipitation projections, which has shown to be the most challenging scenario. Significant improvement was also seen for \emph{winter} and \emph{year} temperature projections.

\begin{figure*}	[htb]
\begin{figrow}
\item \label{rowone} \raisebox{-0.5\height}{
\includegraphics[width=.19\linewidth]{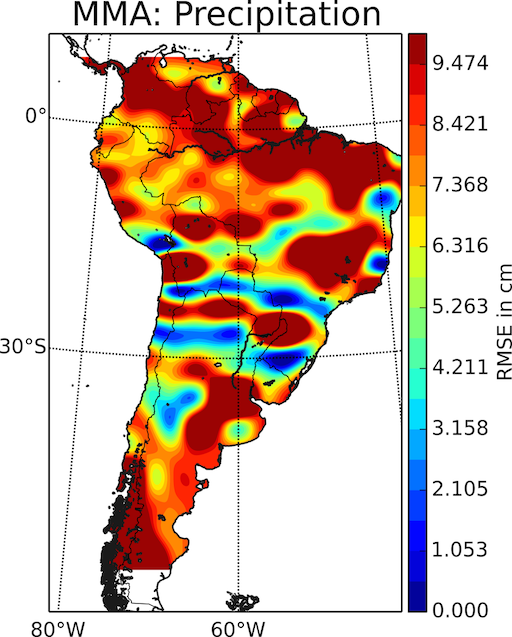} 
\includegraphics[width=.19\linewidth]{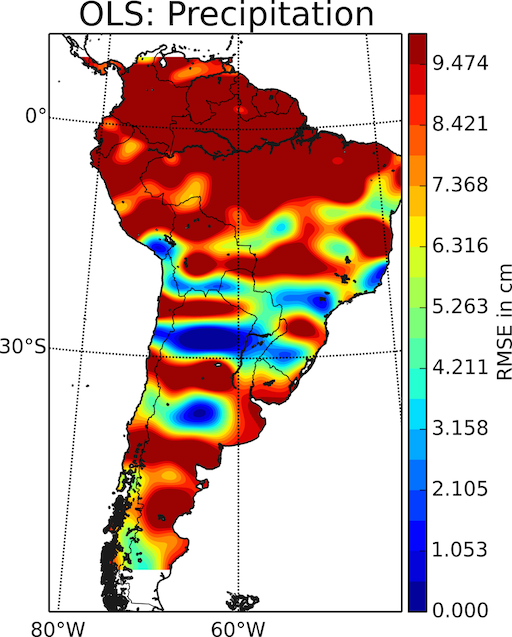}

\includegraphics[width=.19\linewidth]{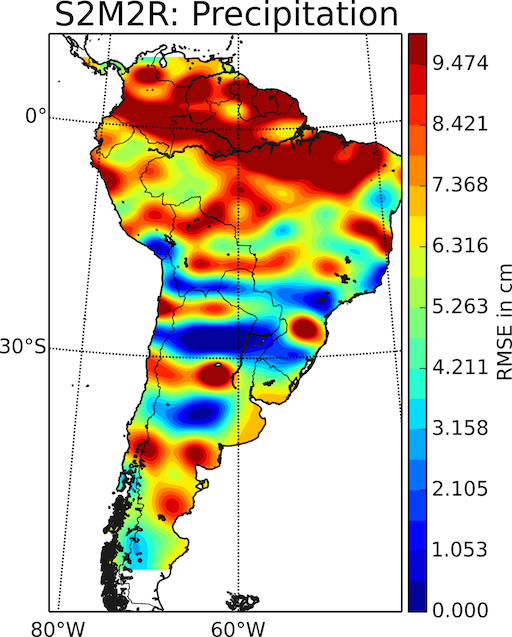}
\includegraphics[width=.19\linewidth]{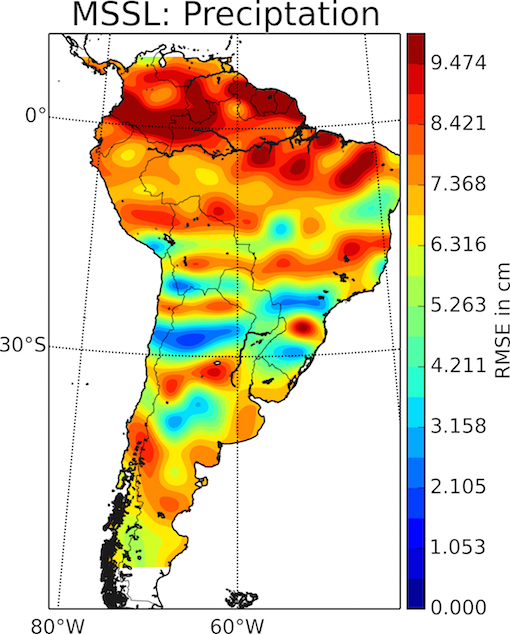}
\includegraphics[width=.19\linewidth]{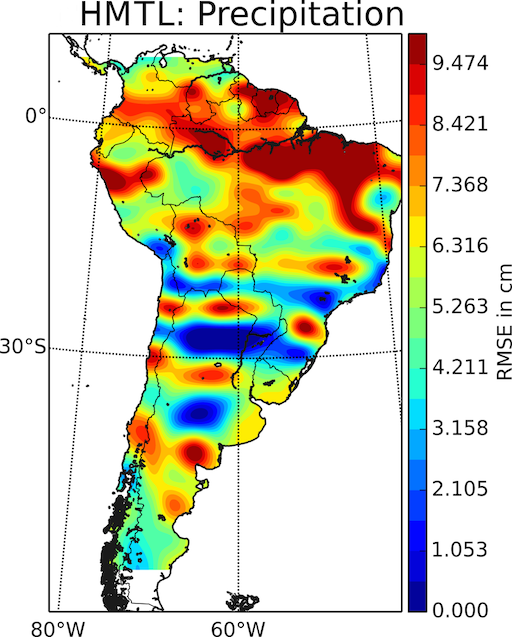}}
\vspace*{10pt}
\item \label{rowtwo}  \raisebox{-0.5\height}{
\includegraphics[width=.19\linewidth]{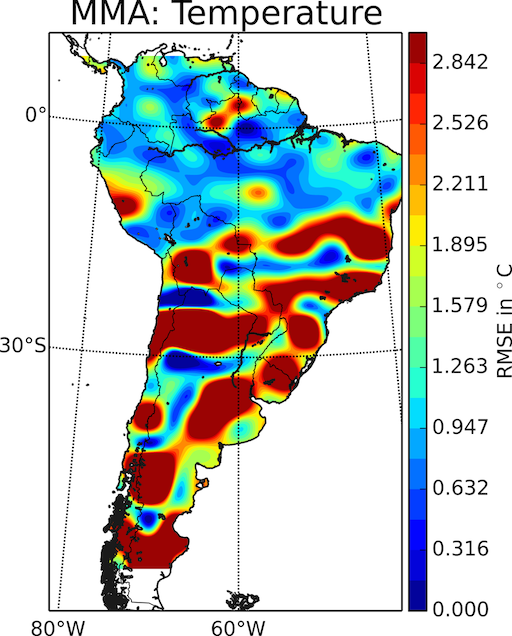}
\includegraphics[width=.19\linewidth]{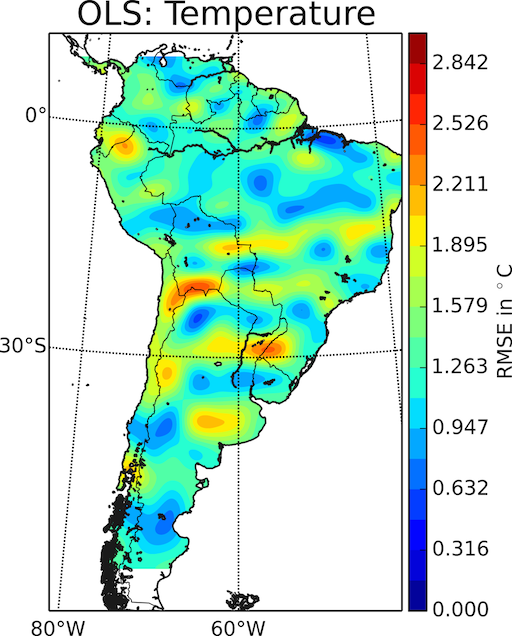}  
\includegraphics[width=.19\linewidth]{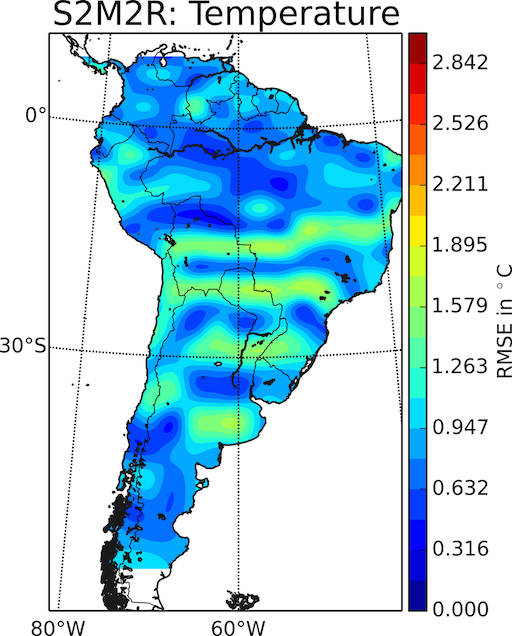}
\includegraphics[width=.19\linewidth]{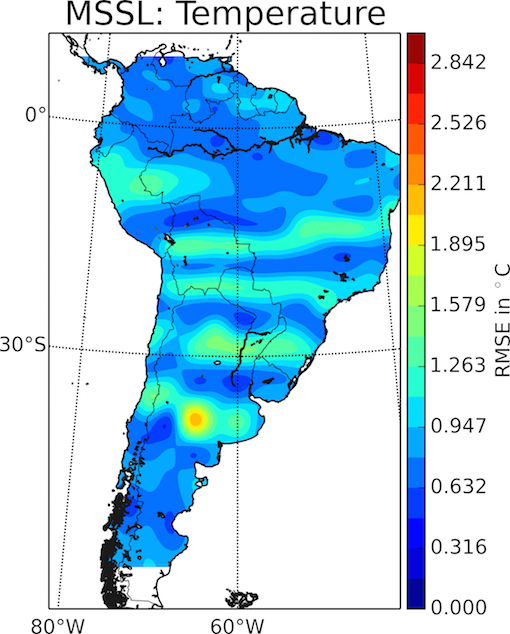}
\includegraphics[width=.19\linewidth]{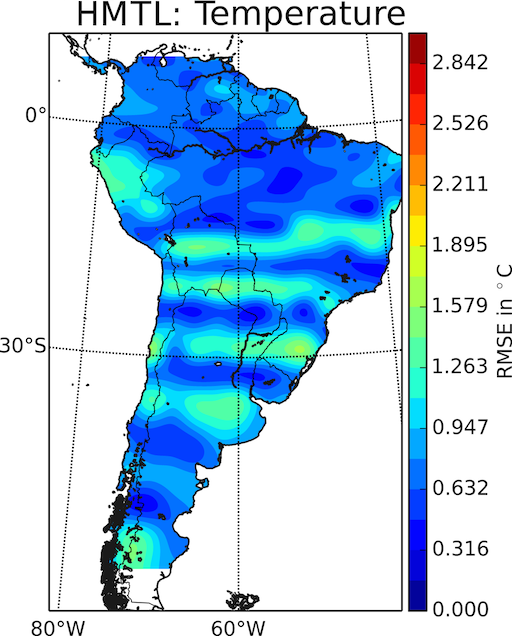}}
\end{figrow}
\caption{Precipitation - row (a), and Temperature - row (b), in summer (20 years): RMSE per geographical location for HMTL and four baselines. HMTL produced more accurate  projections (lower RMSE) than the contenders.}
\label{fig:errormap}
\end{figure*}

	RMSE per geographical location for precipitation and temperature are presented in Figure~\ref{fig:errormap}. For precipitation, more accurate projections (lower RMSE) was obtained by HMTL in Northernmost regions of South America, including Colombia and Venezuela. More accurate temperature projections were obtained in central North region of South America, which comprises part of the Amazon rainforest.
	
	We believe that the improvements of the hierarchical MTL model are due to the same reasons as general MTL models: reduction of sample complexity by leveraging information from other related super-tasks. As a consequence, the precision matrices, which guide information sharing among sub-tasks, are estimated more accurately. The reduced sample complexity probably explains the better climate variable prediction capacity in situations with limited amount of measurements (samples), as the results shown in Table~\ref{tab:relative_gain}.

	\section{Concluding Remarks}

	A hierarchical multitask learning (HMTL) framework to deal with multiple MTL problems is proposed. It was motived by the problem of constructing Earth System Models (ESMs) ensemble for the simultaneous prediction of multiple climate variables. The formulation allows two levels of information sharing: (1) model parameters (coefficients of linear regression); and (2) precision matrices, which encodes the relationship of linear regressors. A group lasso regularization is responsible for capturing similar sparsity patterns across multiple precision matrices. 

	Experiments on joint projection of temperature and precipitation in South America showed that the HMTL produced more accurate predictions in many situations, when compared to the independent execution of existing MTL methods for each climate variable. Simulations on synthetic datasets also showed that the proposed HMTL achieved higher performance as the number of internal MTL problems increase. Here, only temperature and precipitation were used, as they are two of the most studied variables in the climate literature. Future works include a wider analysis with other climate variables, such as geopotential heights and wind directions.
	
	\section{Acknowledgments}
	We thank the anonymous reviewers for their valuable comments. AB was supported by NSF grants IIS-1563950, IIS-1447566, IIS-1447574, IIS-1422557, CCF-1451986, CNS- 1314560, IIS-0953274, IIS-1029711, NASA grant NNX12AQ39A, and gifts from Adobe, IBM, and Yahoo. FJVZ thanks CNPq for the financial support (Process no. 309115/2014-0).
	
	\bibliographystyle{natbib}
	\bibliography{refsall}
	
\end{document}